\documentclass[letterpaper, 10 pt, conference]{ieeeconf}  

\IEEEoverridecommandlockouts 
\usepackage{graphicx} \usepackage{times} \usepackage{amsmath}
\usepackage{amssymb}
\usepackage{algorithmic} \usepackage{algorithm}
\usepackage{epsfig} \usepackage{epstopdf}
\usepackage{multirow} \usepackage{subfigure}
\usepackage{caption} \usepackage{cite}
	\usepackage{flushend} \usepackage{color}	\usepackage{url} 
	\usepackage{gensymb} 
  
\def\R{{\mathbb R}}
  \def\1{{\mathbb I}}

\def\Bx{\mathbf{x}} \def\By{\mathbf{y}}

 \def\BX{\mathbf{X}}

 \def\diag{{\textrm{diag}}}

	\newtheorem{example}{Example}

\renewcommand{\baselinestretch}{0.95} 

\title{
\bf
Uncertainty-Aware Learning from Demonstration Using \\
Mixture Density Networks with Sampling-Free Variance Modeling
}

\author{Sungjoon Choi, Kyungjae Lee, Sungbin Lim, and Songhwai Oh
\thanks{
S. Choi, K. Lee, and S. Oh are with the Department of
Electrical and Computer Engineering and ASRI, 
Seoul National University, Seoul 08826, Korea
(emails: \{sungjoon.choi, kyungjae.lee,
songhwai.oh\}@cpslab.snu.ac.kr).
S. Lim is with the Department of Mathematics,
Korea University (email: sungbin@korea.ac.kr).
}
}
\begin{document}
\maketitle 

\begin{abstract}
In this paper, we propose an uncertainty-aware
learning from demonstration method by presenting
a novel uncertainty estimation method utilizing
a mixture density network appropriate for
modeling complex and noisy human behaviors.
The proposed uncertainty acquisition can be
done with a single forward path without Monte Carlo sampling
and is suitable for real-time robotics applications.
Then, we show that it can be can be decomposed into 
\textit{explained} variance and \textit{unexplained} variance
where the connections between aleatoric and epistemic
uncertainties are addressed.
The properties of the proposed uncertainty measure are
analyzed through three different synthetic examples, 
\textit{absence of data}, \textit{heavy measurement noise}, 
and \textit{composition of functions} scenarios.
We show that each case can be distinguished using 
the proposed uncertainty measure and 
presented 
an uncertainty-aware learning from demonstration method
of an autonomous driving
using this property. 
The proposed uncertainty-aware learning from demonstration
method outperforms other compared methods in terms of
safety using a complex real-world driving dataset.
\end{abstract}
\IEEEpeerreviewmaketitle

\section{Introduction}


Recently, deep learning has been successfully applied
to a diverse range of research areas 
including computer vision \cite{He_16},
natural language processing \cite{Collobert_08}, 
and robotics \cite{Schulman_15}.
When deep networks are kept in a cyber environment
without interacting with an actual physical system,
mis-predictions or malfunctioning of the system
may not cause catastrophic disasters.
However, when using deep learning methods 
in a real physical system involving human beings, 
such as an autonomous car,
safety issues must be considered appropriately \cite{Amodei_16}. 

In May 2016, in fact, a fatal car accident had occurred 
due to the malfunctioning of a low-level image processing component
of an advanced assisted driving system (ADAS)
to discriminate the white side of a trailer from a bright sky
\cite{Tesla_16}.
In this regard, Kendall and Gal \cite{Kendall_17} 
proposed an uncertainty modeling method for deep learning
estimating both 
\textit{aleatoric} and \textit{epistemic} uncertainties indicating
noise inherent in the data generating process
and uncertainty in the predictive model
which captures the ignorance about the model. 
However, computationally-heavy Monte Carlo sampling is
required which makes it not suitable for real-time applications.

In this paper, we present a novel uncertainty estimation method
for a regression task using a deep neural network and its application
to learning from demonstration (LfD).
Specifically, a mixture density network (MDN) \cite{Bishop_94} is used
to model underlying process which is more appropriate
for describing complex distributions \cite{Brando_17}, e.g., 
human demonstrations.
We first present an uncertainty modeling method when
making a prediction with an MDN which can be acquired with 
a single MDN forward path without Monte Carlo sampling. 
This sampling-free property makes it suitable for 
real-time robotic applications compared to 
existing uncertainty modeling methods that require multiple models 
\cite{Lakshminarayanan_16}
or sampling \cite{Gal_16, Gal_16_thesis, Kendall_17}.
Furthermore, as an MDN is appropriate for modeling 
complex distributions \cite{Mclachlan_88} compared to
a density network used in \cite{Kendall_17, Mclachlan_88}
or standard neural network for regression, 
the experimental results on autonomous driving tell us that 
it can better represent the underlying policy of a driver
given complex and noise demonstrations.

The main contributions of this paper are twofold. 
We first present a sampling-free uncertainty estimation method 
utilizing an MDN
and show that it can be decomposed into two parts, 
\textit{explained} and \textit{unexplained} variances
which indicate our ignorance about the model
and measurement noise, respectively. 
The properties of the proposed uncertainty modeling method 
is analyzed through three different cases:
\textit{absence of data}, \textit{heavy measurement noise}, 
and \textit{composition of functions} scenarios. 
Using the analysis, we further propose an uncertainty-aware
learning from demonstration (Lfd) method. 
We first train an aggressive controller in a simulated environment
with an MDN and use the \textit{explained} variance of an MDN
to switch its mode to a rule-based conservative controller.
When applied to a complex real-world driving dataset
from the US Highway $101$ \cite{Colyar_07}, 
the proposed uncertainty-award LfD outperforms compared methods
in terms of safety of the driving as the out-of-distribution inputs, 
which are often refer to as
\textit{covariate shift} \cite{Ross_13}, 
are successfully captured by the proposed \textit{explained} variance.

The remainder of this paper is composed as follows: 
Related work and preliminaries regarding modeling uncertainty
in deep learning are introduced in 
Section \ref{sec:rel} and Section \ref{sec:prel}. 
The proposed uncertainty modeling method with an MDN
is presented in Section \ref{sec:unct} and 
analyzed in Section \ref{sec:toy}.
Finally, in Section \ref{sec:lfd}, we present an uncertainty-aware
learning from demonstration method 
and successfully apply to an autonomous driving task 
using a real-world driving dataset by deploying and controlling
an virtual car inside the road.

\section{Related Work}\label{sec:rel}

Despite the heavy successes in deep learning research areas,
practical methods for estimating uncertainties 
in the predictions with deep networks
have only recently become actively studied.
In the seminal study of Gal and Ghahramani \cite{Gal_16},
a practical method of estimating the predictive variance 
of a deep neural network is proposed by
computing from the sample mean and variance of
stochastic forward paths, i.e., dropout \cite{Srivastava_14}.
The main contribution of \cite{Gal_16} 
is to present a connection between 
an approximate Bayesian network
and a sparse Gaussian process.
This method is often referred to as Monte Carlo (MC) dropout
and successfully applied to modeling model uncertainty in 
regression tasks, classification tasks, 
and reinforcement learning with Thompson sampling. 
The interested readers are referred to \cite{Gal_16_thesis}
for more comprehensive information about uncertainty
in deep learning and Bayesian neural networks.

Whereas \cite{Gal_16} uses a standard neural network, 
\cite{Lakshminarayanan_16} uses
a density network whose output consists 
of both mean and variance of a prediction
trained with a negative log likelihood criterion. 
Adversarial training is also applied by incorporating
artificially generated adversarial examples.
Furthermore, a multiple set of models are trained using
different training sets to form an ensemble where 
the sample variance of a mixture distribution is used
to estimate the uncertainty of a prediction.
Interestingly, the usage of a mixture density network is
encouraged in cases of modeling more complex distributions.
Guillaumes \cite{Brando_17}
compared existing uncertainty acquisition 
methods including \cite{Lakshminarayanan_16, Gal_16}.

Kendall and Gal \cite{Kendall_17} decomposed the
predictive uncertainty into two major types,
\textit{aleatoric} uncertainty and \textit{epistemic} uncertainty. 
First, \textit{epistemic} uncertainty captures 
our ignorance about the predictive model.
It is often referred to as a reducible uncertainty as this type of uncertainty
can be reduced as we collect more training data 
from diverse scenarios.
On the other hand, \textit{aleatoric} uncertainty captures 
irreducible aspects of the predictive variance, such as 
the randomness inherent in the coin flipping. 
To this end, Kendall and Gal utilized a density network similar to 
\cite{Lakshminarayanan_16} but used a slightly different 
cost function for numerical stability. 
The variance outputs directly from the density network
indicates heteroscedastic aleatoric uncertainty where the
overall predictive uncertainty of the output $y$
given an input $\Bx$ is approximated using
\begin{equation} \label{eqn:kendall}
	\mathbb{V}(y|\Bx) \approx \frac{1}{T}
		\sum_{t=1}^{T} \hat{\mu}^2_t(\Bx)
		-\left( 
			\sum_{t=1}^{T} \hat{\mu}_t(\Bx)
		\right)^2
		+\frac{1}{T}\sum_{t=1}^T \hat{\sigma}(\Bx)
\end{equation}
where $\{ \hat{\mu}_t(\Bx), \hat{\sigma}_t(\Bx) \}_{t=1}^T$
are $T$ samples of 
mean and variance functions of a density network
with stochastic forward paths.

Modeling and incorporating uncertainty in predictions
have been widely used in robotics,
mostly to ensure safety 
in the training phase of reinforcement learning
\cite{Kahn_17} 
or to avoid false classification of learned cost function
\cite{Richter_17}.
In \cite{Kahn_17}, an uncertainty-aware collision prediction method
is proposed
by training multiple deep neural networks using bootstrapping and dropout.
Once multiple networks are trained, the sample mean and variance 
of multiple stochastic forward paths of different networks are used
to compute the predictive variance.
Once the predictive variance is higher than a certain threshold, 
a risk-averse cost function is used instead of 
the learned cost function
leading to a low-speed control.
This approach can be seen as extending \cite{Gal_16} by adding 
additional randomness from bootstrapping. 
However, as multiple networks are required,
computational complexities of both training and test phases
are increased. 

In \cite{Richter_17}, safe visual navigation is presented
by training a deep network for modeling 
the probability of collision for a receding horizon control problem.
To handle \textit{out of distribution} cases, a novelty detection
algorithm is presented where the reconstruction loss of
an autoencoder is used as a measure of novelty.
Once current visual input is detected to be novel (high reconstruction loss), 
a rule-based collision estimation is used 
instead of learning based estimation.
This switching between learning based and rule based approaches
is similar to our approach. 
But, we focus on modeling an uncertainty of 
a policy function which consists of both input and output pairs 
whereas the novelty detection in \cite{Richter_17} 
can only consider input data. 

The proposed method can be regarded as extending
previous uncertainty estimation approaches by using 
a mixture density network (MDN) 
and present a novel variance estimation method for an MDN. 
We further show a single MDN without MC sampling
is sufficient to model both reducible and irreducible parts of uncertainty. 
In fact, we show that it can better model both types of uncertainties
in synthetic examples.

\section{Preliminaries} \label{sec:prel}

\subsection{Uncertainty Acquisition in Deep Learning} \label{subsec:unct_dl}

In \cite{Kendall_17}, Kendall and Gal proposed 
two types of uncertainties, 
\textit{aleatoric} and \textit{epistemic} uncertainties.
These two types of uncertainties capture different aspects of
predictive uncertainty, i.e., a measure of uncertainty
when we make a prediction using an approximation method 
such a deep network.
First, \textit{aleatoric} uncertainty captures the uncertainty in the
data generating process, e.g., inherent randomness
of a coin flipping or measurement noise. 
This type of uncertainty cannot be reduced 
even if we collect more training data. 
On the other hand, 
\textit{epistemic} uncertainty models the ignorance of
the predictive model where it can be explained away given 
an enough number of training data.
Readers are referred to Section $6.7$ in \cite{Gal_16_thesis} 
for further details.

Let $\mathcal{D} = \{(\mathbf{x}_{i}, y_{i}) : i = 1, \ldots, n \}$
be a dataset of $n$ samples.
For notational simplicity, we assume that an input and output are
an $d$-dimensional vector, i.e., $\mathbf{x} \in \R^d$, and 
a scalar, i.e., $y \in \R$, respectively. 
Suppose that 
\begin{equation*}
	y=f(\mathbf{x})+\eta
\end{equation*}
where 
$f(\cdot)$ is a target function and 
a measurement error $\eta$ follows a zero-mean Gaussian 
distribution with a variance $\sigma^2_{\eta}$,
i.e., $\eta \sim \mathcal{N}(0, \sigma^2_{\eta})$.
Note that, in this case, 
the variance of $\eta$ corresponds to the \textit{aleatoric} uncertainty, i.e., 
$\sigma^2_{\eta}=\sigma^2_a$ where
we will denote $\sigma^2_a$ as \textit{aleatoric} uncertainty. 
Similarly, \textit{epistemic} uncertainty will be denoted as
$\sigma^2_e$.

Suppose that we train $\hat{f}(\mathbf{x})$ to approximate $f(\mathbf{x})$ 
from $\mathcal{D}$ and get
\begin{equation*}
	\sigma_{e}^{2}
		=\mathbb{E}\left\Vert f(\mathbf{x})-\hat{f}(\mathbf{x})\right\Vert ^{2}.
\end{equation*}
Then, we can see that
\begin{align*}
	\mathbb{E}\left\Vert y-\hat{f}(\mathbf{x})\right\Vert ^{2} 
		& = \mathbb{E}\left\Vert y-f(\mathbf{x})+f(\mathbf{x})-\hat{f}(\mathbf{x})\right\Vert ^{2}
		\\
		&= \mathbb{E}\left\Vert y-f(\mathbf{x}) \right\Vert ^{2}
			+ \mathbb{E}\left\Vert  f(\mathbf{x}) - \hat{f}(\mathbf{x}) \right\Vert ^{2}
		\\
		&=\sigma_{a}^{2}+\sigma_{e}^{2}
\end{align*}
which indicates the total predictive variance is the sum of
\textit{aleatoric} uncertainty and \textit{epistemic} uncertainty.

Correctly acquiring and distinguishing
each type of uncertainty is important to many practical problems. 
Suppose that we are making our model to predict the steering angle 
of an autonomous driving car 
(which is exactly the case in our experiments)
where the training data is collected from human drivers
with an accurate measurement device. 
In this case, 
high \textit{aleatoric} uncertainty and small \textit{epistemic} uncertainty
indicate that there exist multiple possible steering angles. 
For example, a driver can reasonably steer the car 
to both left and right in case of another car in front and both sides open. 
However, when the prediction has 
low \textit{aleatoric} uncertainty and high \textit{epistemic} uncertainty,
this indicates that our model
is uncertain about the current prediction
which could possibly due to the lack of training data.
In this case, it is reasonable to switch to a risk-averse controller
or alarm the driver, if possible.

\subsection{Mixture Density Network} \label{subsec:mdn}

\begin{figure}[!t] \centering
	\includegraphics[width=0.6\columnwidth]{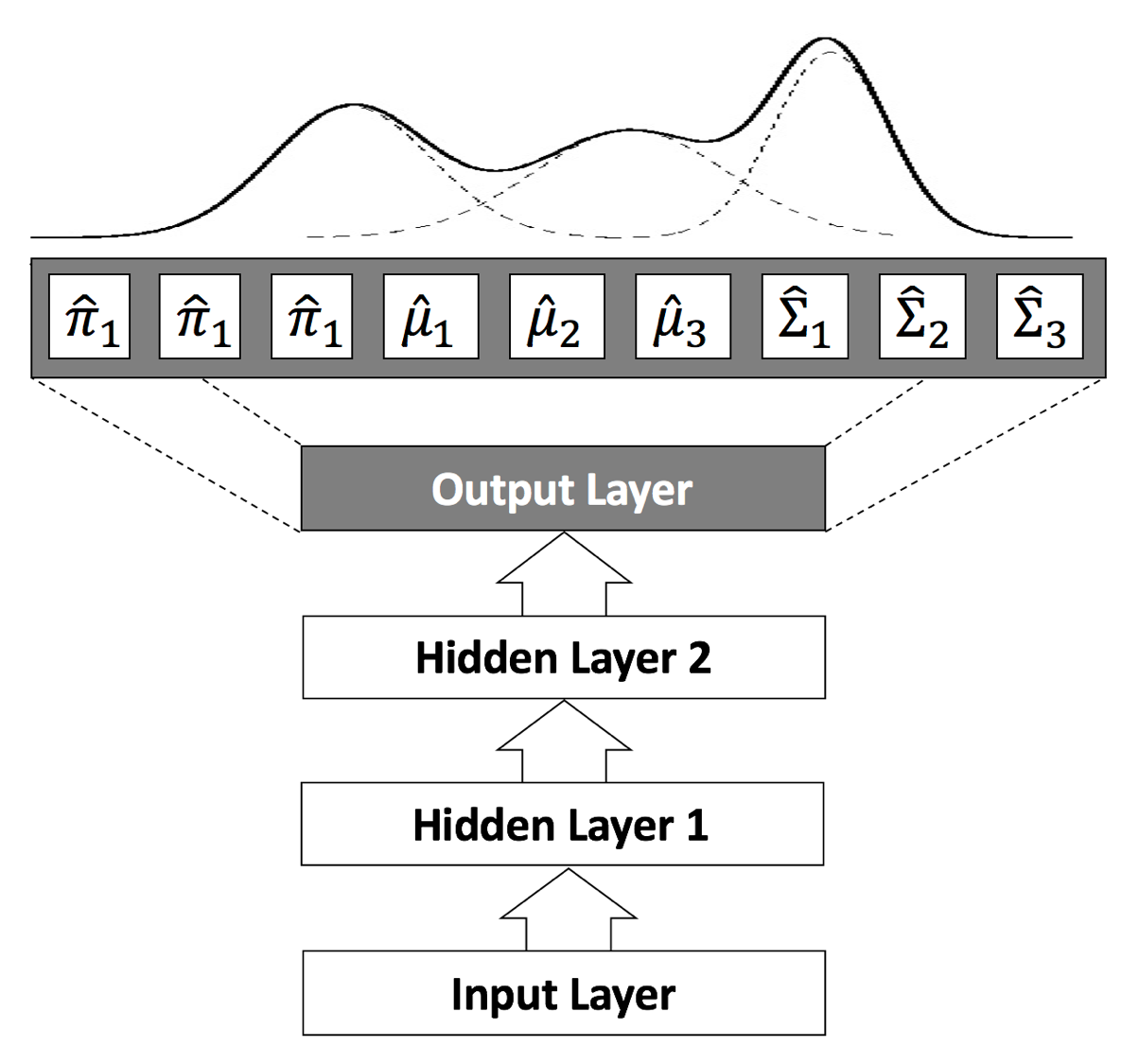}
	\caption{
		A mixture density network ($K=3$) with two hidden layers
		where the output of the network is decomposed into
		$\hat{\pi}$, $\hat{\mu}$, and $\hat{\Sigma}$. 
		}
	\label{fig:mdn}
\end{figure}

A mixture density network (MDN) was first proposed in \cite{Bishop_94}
where the output of a neural network is composed of 
parameters constructing a Gaussian mixture model (GMM):
\begin{equation*}
	p(\By|\theta) 
		= 
		\sum_{j=1}^K \pi_j \mathcal{N}(\By|\mu_j, \, \Sigma_j)
\end{equation*}
where $\theta = \{ \pi_j, \mu_j, \Sigma_j \}_{j=1}^K$ is a set of parameters
of a GMM, mixture probabilities, mixture means, and mixture variances, 
respectively. 
In other words, an MDN can be seen as a mapping $f$ from an input 
$\Bx \in \BX$
to the parameters $\theta \in \Theta$ of a GMM of an output, i.e., 
$f: \BX \mapsto \Theta $ as shown in Figure \ref{fig:mdn}.

However, as mixture weights should be on a $K$ dimensional simplex
and each mixture variance should be positive definite, 
the output of an MDN is handled accordingly.
We assume that each output dimension is independent, 
and each mixture variance becomes a diagonal matrix.
Suppose that the dimension of the output $\By$ is $d$ 
and
let $\{ \hat{\pi}_j, \hat{\mu}_j, \hat{\Sigma}_j \}_{j=1}^K$
be the raw output of an MDN where 
$\hat{\pi}_j \in \R$, $\hat{\mu}_j \in \R^d$, and 
$\hat{\Sigma}_j \in \R^d$. 
Then the parameters of a GMM is computed by
\begin{align}
	\pi_j &= \frac{ \exp \left(
			 \hat{\pi}_j - \max(\pi) 
			 		\right) }
		{\sum_{k=1}^K \exp( \hat{\pi}_k -\max(\pi)  )} 
		\label{eqn:mdn_pi}
		\\
	\mu_j &= \hat{\mu_j} 
		\nonumber
		\\
	\Sigma_j &= \sigma_{max} \diag
		\left(  \rho(\hat{\Sigma}_j)
		\right)
		\label{eqn:mdn_sigma}
\end{align}
where $\max(\pi)$ indicates the maximum mixture weights
among all $K$ weights, 
$\diag(\cdot)$ is an operation to convert a 
$d$-dimensional vector to a $d \times d$-dimensional diagonal matrix, 
and $\rho(x)=\frac{1}{1+\exp(-x)}$ 
is an element-wise sigmoid function.

Two heuristics are applied in (\ref{eqn:mdn_pi})
and (\ref{eqn:mdn_sigma}).
First, 
we found that exponential operations often cause
numerical instabilities, and thus, 
subtracted the maximum mixture value in (\ref{eqn:mdn_pi}). 
In (\ref{eqn:mdn_sigma}), similarly, 
we used a sigmoid function multiplied by a constant
instead of an exponential function 
to satisfy the positiveness constraint of the variance.
$\sigma_{max}$ is selected manually 
and set to five throughout the experiments. 

For training the MDN, we used a negative log likelihood 
as a cost function
\begin{equation}
	c(\theta ; \mathcal{D}) = 
		-\frac{1}{N}
		\sum_{i=1}^N
		\log
		(
			\sum_{j=1}^K \pi_j(\Bx_i) 
				\mathcal{N}(\By_i | \mu_j(\Bx_i), \, \Sigma_j(\Bx_i))
			+ \epsilon
		)
\end{equation}
where $\mathcal{D} = \{(\Bx_{i}, \By_{i}) \}_{i=1}^N$
is a set of training data and $\epsilon=10^{-6}$ is for numerical 
stability of a logarithm function. 
We would like to note that an MDN can be implemented 
on top of any deep neural network architectures, 
e.g., a multi-layer perceptron, convolutional neural network, 
or recurrent neural network. 

Once an MDN is trained, the predictive mean and variance 
can be computed by selecting the mean and variance of the
mixture of the highest mixture weight (MAP estimation).
This can be seen as a mixture of experts \cite{Shazeer_17}
where the mixture weights form a gating network for 
selecting local experts. 

\section{Proposed Uncertainty Estimation Method} \label{sec:unct}

In this section, we propose a novel uncertainty acquisition method
for a regression task using a mixture density network (MDN)
using a law of total variance \cite{Duda_73}.
As described in Section \ref{subsec:mdn}, an MDN 
constitutes a Gaussian mixture model (GMM) 
of an output $\mathbf{y}$ 
given a test input $\mathbf{x}$:
\begin{equation} \label{eq:GMM}
	p(\mathbf{y}|\mathbf{x})
		= \sum_{j=1}^{K}\pi_{j}(\mathbf{x})
		\mathcal{N}\left(
			\mathbf{y};\mu_{j}(\mathbf{x}), \, \Sigma_{j}(\mathbf{x})
			\right)
\end{equation}
where $\pi_j(\mathbf{x})$, $\mu_j(\mathbf{x})$, and $ \Sigma_{j}(\mathbf{x}) $
are $j$-th mixture weight function, mean function, and variance function, 
respectively. 
Note that a GMM can approximate any given density function 
to arbitrary accuracy \cite{Mclachlan_88}.
While the number of mixtures becomes prohibitively large
to achieve this, it is worthwhile noting that an MDN is more suitable for fitting
complex and noisy distribution compared to a density network.

\subsection{Uncertainty Acquisition for a Mixture Density Network}

Let us first define the total expectation of a GMM. 
\begin{align}
	\mathbb{E}[\mathbf{y}|\mathbf{x}] 
		& = \sum_{j=1}^{K}\pi_{j}(\mathbf{x}) 
			\int\mathbf{y}\mathcal{N}(\mathbf{y};\mu_{j}(\mathbf{x}), 
				\Sigma_{j}(\mathbf{x}))d\mathbf{y} \nonumber \\
		& = \sum_{j=1}^{K}\pi_{j}(\mathbf{x})\mu_{j}(\mathbf{x})
\end{align} \label{eq:condi_exp}
The total variance of a GMM is computed as follows
(we omit $\mathbf{x}$ in each function).
\begin{align*}
	\mathbb{V}(\mathbf{y}|\mathbf{x})  
		& = \int\left\Vert \mathbf{y}-\mathbb{E}\left[\mathbf{y}|\mathbf{x}\right]\right\Vert ^{2}
			p(\mathbf{y}|\mathbf{x})d\mathbf{y}
		\\
		& = \sum_{j=1}^{K}\pi_{j}\int\left\Vert 
			\mathbf{y}-\sum_{k=1}^{K}\pi_{k}\mu_{k}\right\Vert^{2}
			\mathcal{N}(\mathbf{y};\mu_{j},\Sigma_{j})d\mathbf{y}
\end{align*}
where the term inside the integral becomes
\begin{align*}
	\int & \left\Vert \mathbf{y}-\sum_{k=1}^{K}\pi_{k}\mu_{k}\right\Vert^{2}
		\mathcal{N}(\mathbf{y};\mu_{j},\Sigma_{j})d\mathbf{y}
		\\
	& = \int\left\Vert \mathbf{y}-\mu_{j}\right\Vert ^{2}
		\mathcal{N}(\mathbf{y};\mu_{j},\Sigma_{j})d\mathbf{y}
	\\
	& \qquad+\int\left\Vert \mu_{j}-\sum_{k=1}^{K}\pi_{k}\mu_{k}\right\Vert ^{2}
		\mathcal{N}(\mathbf{y};\mu_{j},\Sigma_{j})d\mathbf{y}
	\\
	& \qquad\quad+ 2 \int(\mathbf{y}-\mu_{j})^{T}(\mu_{j}-
		\sum_{k=1}^{K}\pi_{k}\mu_{k})
		\mathcal{N}(\mathbf{y};\mu_{j},\Sigma_{j})d\mathbf{y}
	\\
	& = \Sigma_{j}+\left\Vert \mu_{j}-\sum_{k=1}^{K}\pi_{k}\mu_{k}\right\Vert ^{2}.
\end{align*}
Therefore the total variance $\mathbb{V}(\mathbf{y}|\mathbf{x})$
becomes
(also see (47) in \cite{Bishop_94}):
\begin{align} 
	\mathbb{V}(\mathbf{y}|\mathbf{x})
		&= \sum_{j=1}^{K}\pi_{j}(\mathbf{x})\Sigma_{j}(\mathbf{x}) 
		\nonumber
		\\ 
		&\quad+\sum_{j=1}^{K}\pi_{j}(\mathbf{x}) 
			\left \Vert \mu_{j}(\mathbf{x})-\sum_{k=1}^{K}\pi_{k}(\mathbf{x})
				\mu_{k}(\mathbf{x})\right\Vert ^{2}.
		\label{eq:condi_var}
\end{align}
Now let us present the connection between
the two terms in (\ref{eq:condi_var})
that constitute the total variance of a GMM
to the \textit{epsitemic} uncertainty
and \textit{aleatoric} uncertainty in \cite{Kendall_17}.

\begin{figure*}[t] \centering
	\subfigure[]{\includegraphics[width=0.4\columnwidth]
	{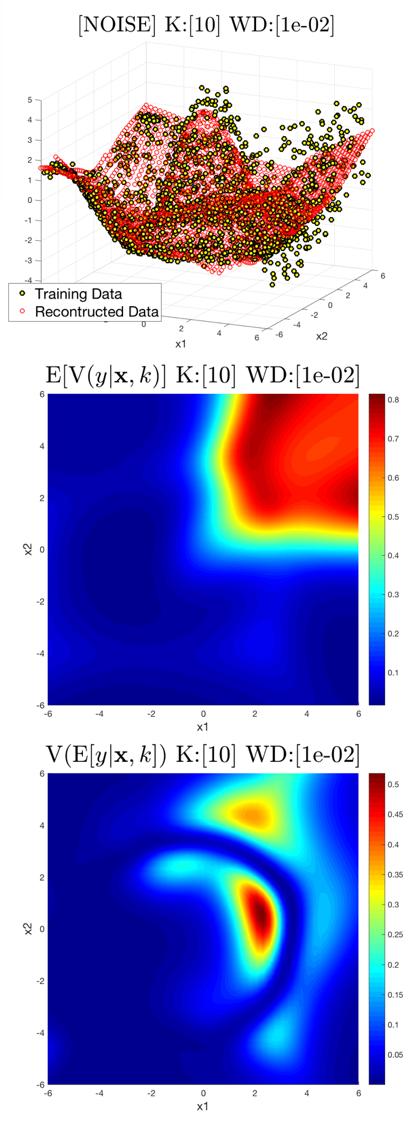} 
	\label{fig:noise_k10}}
	\subfigure[]{\includegraphics[width=0.4\columnwidth]
	{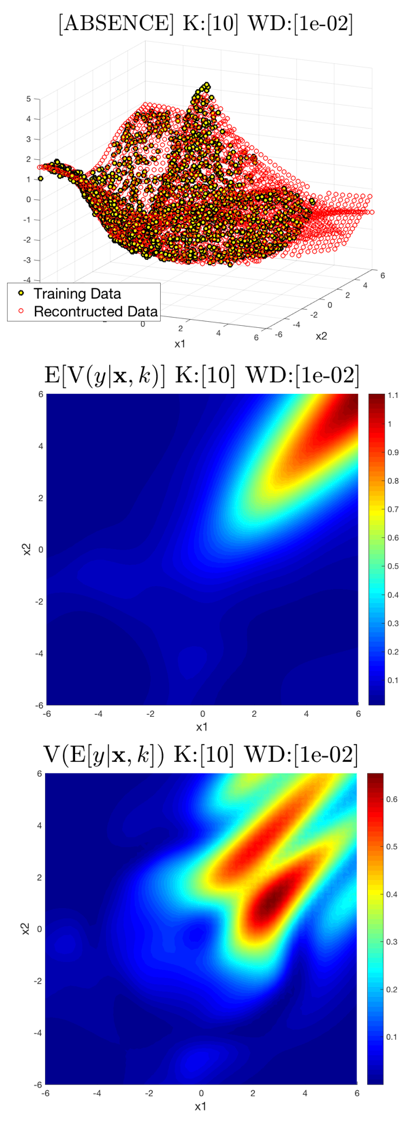} 
	\label{fig:absence_k10}}
	\subfigure[]{\includegraphics[width=0.4\columnwidth]
	{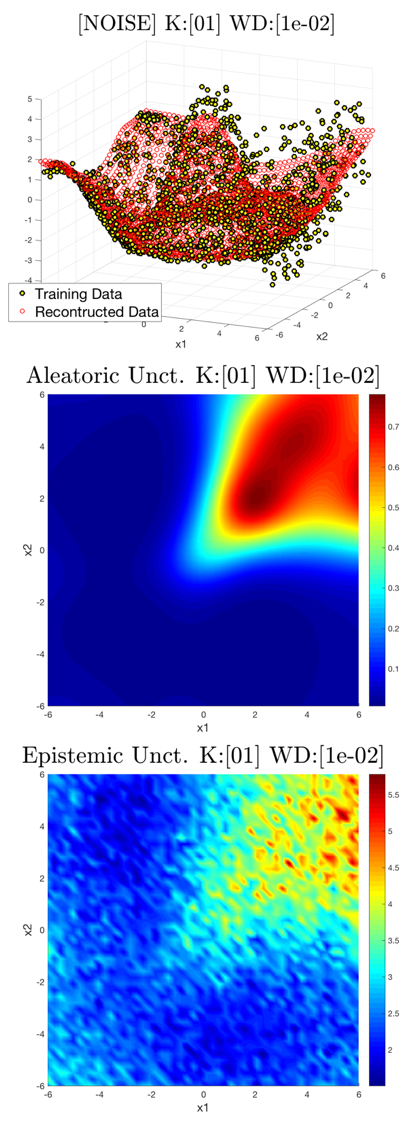} 
	\label{fig:noise_k01}} 
	\subfigure[]{\includegraphics[width=0.4\columnwidth]
	{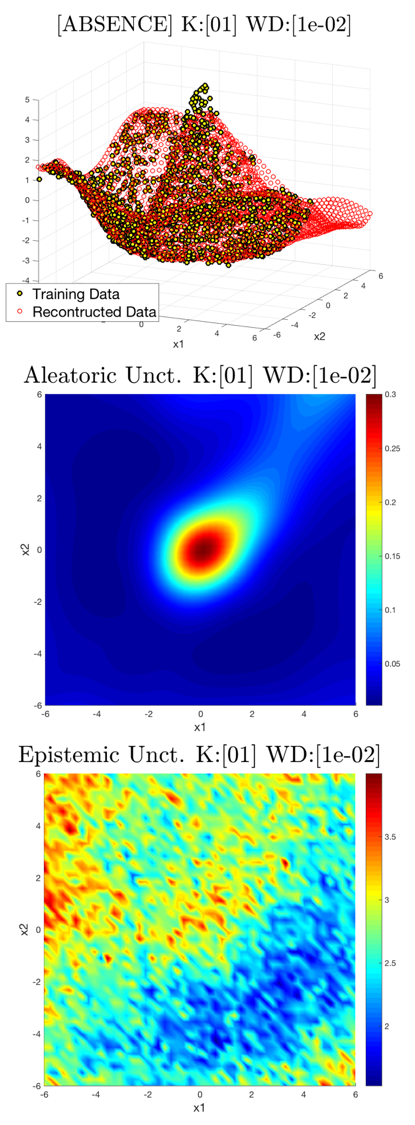} 
	\label{fig:absence_k01}} 
	\caption{
		Proposed uncertainty measures of a) \textit{heavy noise}
		and b) \textit{absence of data} scenarios and
		\textit{Aleatoric} and \textit{epistemic} uncertainties
		of (c) \textit{heavy noise} and (d) \textit{absence of data} scenarios. 
	}
	\label{fig:noise_absence}
\end{figure*}

\subsection{Connection to Aleatoric and Epistemic Uncertainties}

Let $\sigma_{\mathfrak{a}}$ and 
$\sigma_{\mathfrak{e}}$ be \textit{aleatoric} uncertainty 
and \textit{epistemic} uncertainty, respectively. 
Then, these two uncertainties constitute total predictive variance
as shown in Section \ref{subsec:unct_dl}. 
\[
	\mathbb{V}(\mathbf{y}|\mathbf{x})
		=\sigma_{\mathfrak{a}}^{2}(\mathbf{x})
		+\sigma_{\mathfrak{e}}^{2}(\mathbf{x})
\]
On the other hand, we can rewrite (\ref{eq:condi_var}) as
\begin{align*}
	\mathbb{V}\left(\mathbf{y}|\mathbf{x}\right) 
		& = \mathbb{V}_{k\sim\pi}
			\left(\mathbb{E}\left[\mathbf{y}|\mathbf{x},k\right]\right)
		+\mathbb{E}_{k\sim\pi}
			\left(\mathbb{V}\left(\mathbf{y}|\mathbf{x},k\right)\right).
\end{align*}
We remark that 
$\mathbb{V}_{k\sim\pi}(\mathbb{E}[\mathbf{y}|\mathbf{x},k])$
indicates \emph{explained} variance 
whereas $\mathbb{E}_{k\sim\pi}\left[\mathbb{V}(\mathbf{y}|\mathbf{x},k)\right]$
represents \emph{unexplained} variance \cite{Duda_73}.
Observe that
\begin{align} 
	&
	\mathbb{V}_{k\sim\pi}(\mathbb{E}[\mathbf{y}|\mathbf{x},k])
	= 
	\mathbb{V}_{k\sim\pi}\left(\mu_{k}(\mathbf{x})\right) 
	\nonumber
	\\
	 &=
	 \sum_{k=1}^{K}\pi_{k}(\mathbf{x})
		\left\Vert 
			\mu_{k}(\mathbf{x})-\sum_{j=1}^{K}\pi_{j}(\mathbf{x})\mu_{j}(\mathbf{x})
		\right\Vert ^{2} 
	 \label{eqn:ve}	
\end{align} 
\begin{equation}
	\mathbb{E}_{k\sim\pi}[\mathbb{V}(\mathbf{y}|\mathbf{x},k)]
		=
		\mathbb{E}_{k\sim\pi} \left[\Sigma_{k}(\Bx) \right]
		=
		\sum_{k=1}^{K}\pi_{k}(\mathbf{x})\Sigma_{k}(\mathbf{x})
	\label{eqn:ev}	
\end{equation}
This implies that \eqref{eq:condi_var} can be decomposed into uncertainty
quantity of each mixture, i.e., 
\begin{align} 
	\sigma_{\mathfrak{a}}^{2}(\mathbf{x}|k)
	&+
	\sigma_{\mathfrak{e}}^{2}(\mathbf{x}|k)
	\nonumber
	\\ 
	&=
	\Sigma_{k}(\mathbf{x})
	+
		\left\Vert \mu_{k}(\mathbf{x})
		-\sum_{j=1}^{K}\pi_{j}(\mathbf{x})\mu_{j}(\mathbf{x})\right\Vert ^{2}.
	\label{eqn:unct_mixture}
\end{align}
$\Sigma_k(\Bx)$ in the right hand side of (\ref{eqn:unct_mixture})
is the predicted variance of the $k$-th mixture 
where it can be interpreted as \textit{aleatoric} uncertainty
as the variance of a density network captures the noise
inherent in data \cite{Kahn_17}. 
Consequently, 
$
\left\Vert \mu_{k}(\mathbf{x})
		-\sum_{j=1}^{K}\pi_{j}(\mathbf{x})\mu_{j}(\mathbf{x})\right\Vert ^{2}
$
corresponds to the \textit{epistemic} uncertainty estimating 
our ignorance about the model prediction. 
We validate these connections
with both synthetic examples and track driving demonstrations
in Section \ref{sec:toy} and \ref{sec:lfd}. 

We would like to emphasize that
Monte Carlo (MC) sampling is not required 
to compute the total variance of a GMM 
as we introduced randomness from the mixture distribution $\pi$. 
The predictive variance (\ref{eqn:kendall}) 
proposed in in \cite{Kendall_17}
requires MC sampling with random weight masking.
This additional sampling is required as
the density network can only model a single mean and variance.
However, when using the MDN,
it can not only model the measurement noise
or \textit{aleatoric} uncertainty $\sigma^2_a$ with (\ref{eqn:ev})
but also model our ignorance about the model
through (\ref{eqn:ve}).
Intuitively speaking, (\ref{eqn:ve}) becomes high
when the mean function of each mixture does not match 
the total expectation
and will be used to model 
in uncertainty-aware learning from demonstration
in Section \ref{sec:lfd}.

\section{Analysis of the Proposed Uncertainty Modeling with Synthetic Examples} \label{sec:toy}

In this section, we analyze the properties of 
the proposed uncertainty modeling method in Section \ref{sec:unct}
with three carefully designed synthetic examples: 
\textit{absence of data}, \textit{heavy noise}, and 
\textit{composition of functions} scenarios.
In all experiments, fully connected networks
with two hidden layers, $256$ hidden nodes, and 
\textit{tanh} activation function\footnote{
Other activation functions such as \textit{relu} or \textit{softplus}
had been tested as well but omitted as they showed poor regression 
performance on our problem. 
We also observe that increasing the number of mixtures over $10$
does not affect the overall prediction and uncertainty estimation performance.}
is used and 
the number of mixtures is $10$. 
We also implemented the algorithm in \cite{Kendall_17} with the 
same network topology and used $50$ Monte Carlo (MC)
sampling for computing the predictive variance
as shown in the original paper.
The keep probability of dropout is $0.8$ and 
all codes are implemented with TensorFlow \cite{Abadi_16}

In the \textit{absence of data} scenario, we removed the training data 
on the first quadrant and show how different 
each type of uncertainty measure behaves on the input regions 
with and without training data.
For the \textit{heavy noise} scenario, we added heavy uniform noises from $-2$ to $+2$
to the outputs training data whose inputs are on the first quadrant.
The input space and underlying function for both \textit{absence of data}
and \textit{heavy noise} are
$f(\Bx) = 5\cos(\frac{\pi}{2} \|\frac{\Bx}{2}\| )\exp(-\frac{\pi \| \Bx \| }{20})$
and $\mathbb{X}= \{ \Bx=(x_1, x_2) | -6 \le x_1, x_2 \le +6\} $, respectively. 
These two scenarios are designed to validate whether the proposed method
can distinguish unfamiliar inputs from inputs with high measurement errors.
We believe this ability of \textit{knowing its ignorance} is especially important
when it comes to deploy a learning-based system to 
an actual physical system involving humans.

\begin{figure*}[t] \centering
	\subfigure[]{\includegraphics[width=0.38\columnwidth]
	{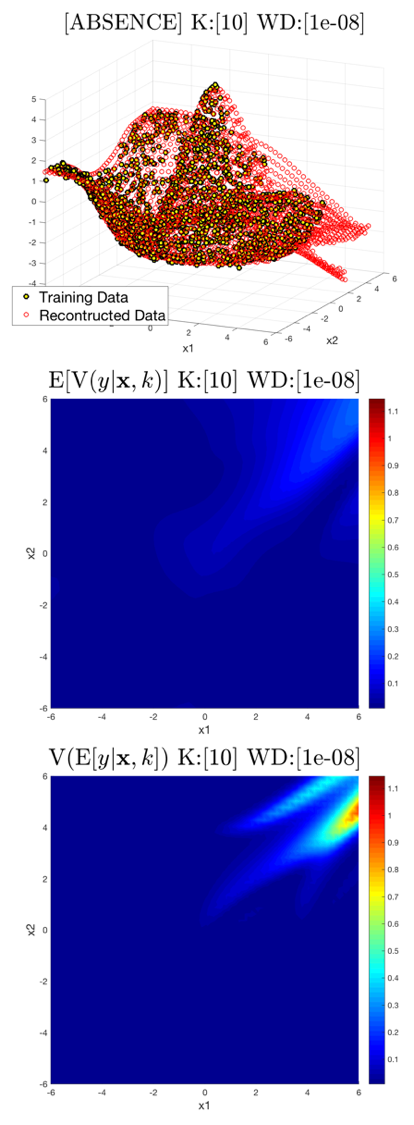} 
	\label{fig:absence_k10_wd1e-8}}
	\subfigure[]{\includegraphics[width=0.38\columnwidth]
	{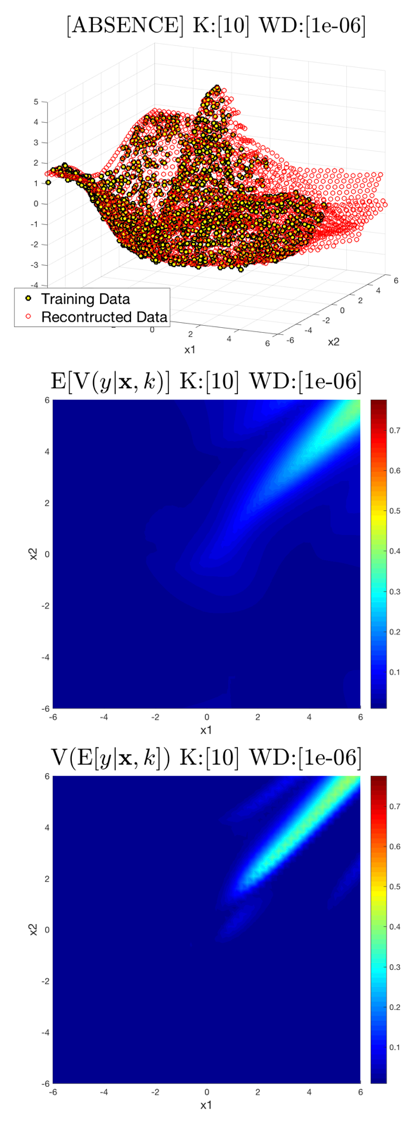} 
	\label{fig:absence_k10_wd1e-6}} 
	\subfigure[]{\includegraphics[width=0.38\columnwidth]
	{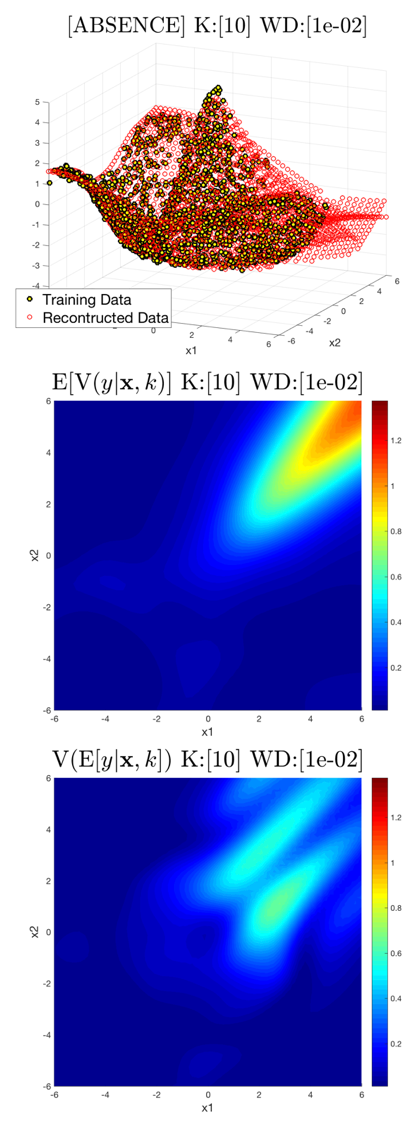} 
	\label{fig:absence_k10_wd1e-2}}
	\subfigure[]{\includegraphics[width=0.38\columnwidth]
	{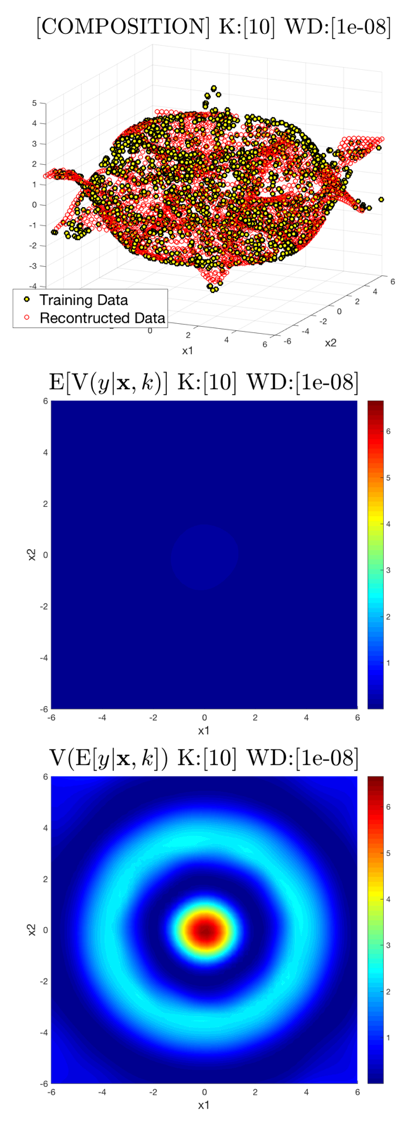} 
	\label{fig:composition_k10}} 
	\subfigure[]{\includegraphics[width=0.38\columnwidth]
	{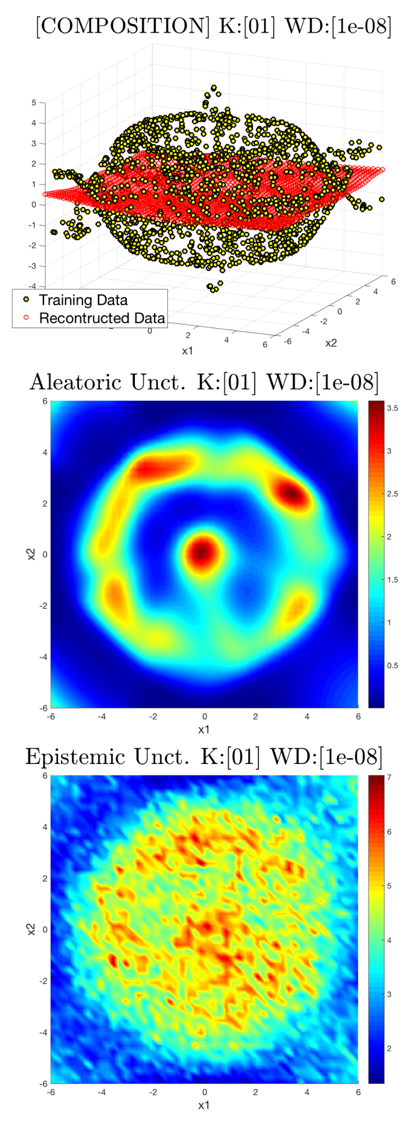} 
	\label{fig:composition_k01}} 
	\caption{
		Proposed uncertainty measures of 
		the \textit{heavy noise} scenario
		while varying the weight decay levels (a-c) and 
		the \textit{composition of functions} scenario (d). 
		(e) \textit{Aleatoric} and \textit{epistemic} uncertainties of
		\textit{composition of functions} scenario.
	}
	\label{fig:tracktrajres}
\end{figure*}

In the \textit{composition of functions} scenario, the training data
are collected from two different function $f(\Bx)$ and flipped $-f(\Bx)$. 
This scenario is designed to reflect the cases where we have 
multiple but few choices to select. 
For example, when there is a car in front, there could be roughly 
two choices to select, turn left or turn right, where both 
choices are totally fine. 
If the learned model can distinguish this from noisy measurements,
it could give additional information to the overall system.

As explained in Section \ref{sec:unct}, the proposed uncertainty measure
is composed of two terms 
$\mathrm{V}(\mathrm{E}[y|\Bx, k])$
and $\mathrm{E}[\mathrm{V}(y|\Bx, k)]$ 
where the sum is the total predictive variance
$\mathrm{V}(y|\Bx)$.
Before starting the analysis,  
let us state again the connections between
$\mathrm{V}(\mathrm{E}[y|\Bx, k])$ and 
$\mathrm{E}[\mathrm{V}(y|\Bx, k)]$ of the proposed method and 
\textit{aleatoric} and \textit{epistemic} uncertainties
proposed in \cite{Kendall_17}. 
$\mathrm{V}(\mathrm{E}[y|\Bx, k])$ indicates 
\textit{explained} variance which
corresponds to \textit{epistemic} uncertainty
and it indicate our ignorance about the model.
On the other hand, 
$\mathrm{E}[\mathrm{V}(y|\Bx, k)]$ models
\textit{unexplained} variance and it corresponds to
\textit{aleatoric} uncertainty indicating measurement noises
or randomness inherent in the data generating process.

In Figure \ref{fig:noise_k10} and \ref{fig:absence_k10}, 
the proposed uncertainty measures, 
$\mathrm{V}(\mathrm{E}[y|\Bx, k])$, $\mathrm{E}[\mathrm{V}(y|\Bx, k)]$, 
and $\mathrm{V}(y|\Bx)$ of both 
\textit{heavy noise} and \textit{absence of data}
scenarios are illustrated. 
The uncertainty measures in \cite{Kendall_17},
\textit{epistemic} uncertainty and \textit{aleatoric} uncertainty,
are shown in Figure \ref{fig:noise_k01} and \ref{fig:absence_k01}.

First, in the \textit{heavy noise} scenario, both proposed method and 
the method in \cite{Kendall_17} capture noisy regions as
illustrated in Figure \ref{fig:noise_k10} and \ref{fig:noise_k01}.
Particularly,  
$\mathrm{E}[\mathrm{V}(y|\Bx, k)]$ in 
Figure \ref{fig:noise_k10}
and
\textit{epistemic} uncertainty in 
Figure \ref{fig:noise_k01}
correctly depicts the region with heavy noise.
This is mainly because measurement noise is related to
the \textit{aleatoric} uncertainty which could easily be captured
with density networks.

\begin{table}[h!]  \center 
	\begin{tabular}{ l | c | c  } 
                    & $\mathrm{E}[\mathrm{V}(y|\Bx, k)]$ & $\mathrm{V}(\mathrm{E}[y|\Bx, k])$ \\
		\hline \hline
		\textit{Heavy noise}	
			& High 		& High or Low
		\\ 
		\textit{Absence of data}     
			& High 		& High			
		\\
		\textit{Composition of functions}  	
			& Low 		& High
		\\
	\end{tabular}
	\caption{Summary of 
		$\mathrm{E}[\mathrm{V}(y|\Bx, k)]$
		and 
		$\mathrm{V}(\mathrm{E}[y|\Bx, k])$ on 
		\textit{absence of data}, \textit{heavy noise}, and 
		\textit{composition of functions} scenarios.
	}
	\label{tbl:summ}
	\begin{tabular}{ l | c | c  } 
                    & Computation Time [ms]
                  \\
		\hline \hline
		\textit{Proposed method}	
			& $48.08$ 	
		\\ 
		\textit{\cite{Kendall_17}}     
			& $1209.12$ 	
		\\
	\end{tabular}
	\caption{Summary of 
		Computation time for estimating 
		the proposed 
		$\mathrm{E}[\mathrm{V}(y|\Bx, k)]$ 
		and 
		$\mathrm{V}(\mathrm{E}[y|\Bx, k])$
		vs.
		\textit{aleatoric} and \textit{epistemic}
		uncertainties in 
		\cite{Kendall_17} where we run the MC sampling
		for $50$ times. 
	}
	\label{tbl:time}
\end{table}

However, when it comes to the \textit{absence of data} scenario,
proposed and compared methods show clear difference.  
While both \textit{aleatoric} and \textit{epistemic} uncertainties 
in Figure \ref{fig:absence_k01}
can hardly capture the regions with no training data,
the proposed method 
shown in Figure \ref{fig:absence_k10}
effectively captures such regions by assigning 
high uncertainties to unseen regions. 
In particular, it can be seen that $\mathrm{V}(\mathrm{E}[y|\Bx, k])$ 
captures this region more suitably compared to 
$\mathrm{E}[\mathrm{V}(y|\Bx, k)]$. 
This is a reasonable result in that $\mathrm{V}(\mathrm{E}[y|\Bx, k])$
is related to the \textit{epistemic} uncertainty 
which represents the model uncertainty,
i.e., our ignorance about the model.

This difference between $\mathrm{V}(\mathrm{E}[y|\Bx, k])$ 
and $\mathrm{E}[\mathrm{V}(y|\Bx, k)]$ becomes more clear when 
we compare
$\mathrm{V}(\mathrm{E}[y|\Bx, k])$ of \textit{absence of data}
and \textit{heavy noise} scenarios.
Unlike \textit{absence of data} scenario where 
$\mathrm{V}(\mathrm{E}[y|\Bx, k])$ is high in the first quadrant
(data-absence region), 
$\mathrm{V}(\mathrm{E}[y|\Bx, k])$ in this region
contains both high and low variances.
This is mainly due to the fact that $\mathrm{V}(\mathrm{E}[y|\Bx, k])$
is related to the \textit{epistemic} uncertainty which can be 
explained away with more training data
(even with high measurement noise).

It is also interesting to see the effect of the prior information 
of weight matrices. 
In Bayesian deep learning, a prior distribution is given to the weight matrices
to assign posterior probability distribution over the outputs
\cite{Gal_16_thesis}.
In this perspective, $L_2$ weight decay 
on the weight matrices can be seen as assigning a Gaussian 
prior over the weight matrices. 
Figure \ref{fig:absence_k10_wd1e-8}, \ref{fig:absence_k10_wd1e-6},
and \ref{fig:absence_k10_wd1e-2} show
$\mathrm{V}(\mathrm{E}[y|\Bx, k])$, 
$\mathrm{E}[\mathrm{V}(y|\Bx, k)]$,
and
$\mathrm{V}(y|\Bx)$
of the 
\textit{absence of data} scenario
with different weight decay levels.
One can clearly see that the regions with no training data
are more accurately captured 
as we increase the weight decay level.
Specifically, $\mathrm{E}[\mathrm{V}(y|\Bx, k)]$
is more affected by the weight decay level
as it corresponds to \textit{aleatoric} uncertainty
which is related to the weight decay level \cite{Gal_16_thesis}.
Readers are referred to Section 6.7 in \cite{Gal_16_thesis}
for more information about the effects of weight decay and
dropout to a Bayesian neural network. 

\begin{figure*}[!t] \centering
	\includegraphics[width=1.9\columnwidth]{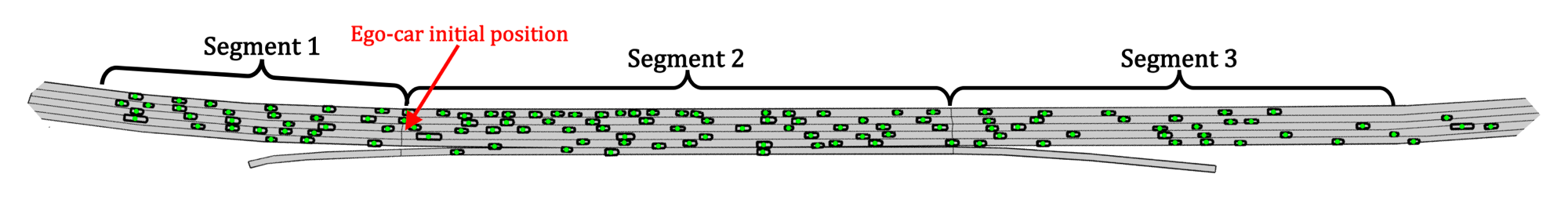}
	\label{fig:trackres_a}
	\caption{
		A snapshot of NGSIM track environmet. 
		}
	\label{fig:ngsim}
\end{figure*}

Figure \ref{fig:composition_k10} and \ref{fig:composition_k01}
shows the experimental results in the
\textit{composition of functions} scenario. 
As a single density network cannot model the composite of two 
functions, the reconstructed results shown with red circles 
in Figure \ref{fig:composition_k01} are poor as well as 
\textit{aleatoric} and \textit{epistemic} uncertainties. 
On the other hand, the reconstructed results from 
the MDN with $10$ mixtures 
accurately model the composition of two function. \footnote{
A composite of two functions is not a proper function as
there exist two different outputs per a single input. 
However, an MDN can model this as it
consists of multiple mean functions. 
}
Furthermore, 
$\mathrm{E}[\mathrm{V}(y|\Bx, k)]$
and $\mathrm{V}(\mathrm{E}[y|\Bx, k])$
show clear differences.
In particular, 
$\mathrm{E}[\mathrm{V}(y|\Bx, k)]$
is low on almost everywhere whereas 
$\mathrm{V}(\mathrm{E}[y|\Bx, k])$
has both high and low variances
which is proportional to the difference
between two composite function. 
As the training data itself does not contain 
any measurement noises,
$\mathrm{E}[\mathrm{V}(y|\Bx, k)]$
has low values in its input domain. 
However, 
$\mathrm{V}(\mathrm{E}[y|\Bx, k])$ becomes high 
where the differences between two possible outputs
are high, as it becomes more hard to fit the training data
in such regions. 

Table \ref{tbl:summ} summarizes how 
$\mathrm{E}[\mathrm{V}(y|\Bx, k)]$
and 
$\mathrm{V}(\mathrm{E}[y|\Bx, k])$
behave on three different scenarios. 
The computation times for the proposed method
and compared method \cite{Kendall_17} is shown in 
Table \ref{tbl:time} where the proposed method
is about $21.15$ times faster as it does not 
require MC sampling.

\section{Uncertainty-Aware Learning from Demonstration to Drive} \label{sec:lfd}


We propose uncertainty-aware LfD (UALfD) which
combines the learning-based approach with
a rule-based approach by switching the mode of the controller 
using the uncertainty measure in Section \ref{sec:toy}. 
In particular, 
\textit{explained} variance (\ref{eqn:ve}) is used as a measure
of uncertainty 
as it estimates the model uncertainty.
The proposed method makes the best of both approaches
by using the model uncertainty as a switching criterion.
The proposed UALfD is applied to an aggressive driving task
using a real-world driving dataset \cite{Colyar_07}
where the proposed method significantly improves the 
performance of driving in terms of both safety and efficiency
by incorporating the uncertainty information. 

For evaluating the proposed uncertainty-aware learning from demonstration
method, we use the Next-Generation Simulation (NGSIM)
datasets collected from
US Highway 101 \cite{Colyar_07} which provides
$45$ minutes of real-world vehicle trajectories at $10$Hz
as well as CAD files for road descriptions. 
Figure \ref{fig:ngsim} illustrates the road configuration of 
US Highway 101, which consists of three segments
and six lanes. 
For testing, we only used the second segment, where the initial position 
of an ego car is at the start location of the third lane in the second segment
and the goal is to reach the third segment. 
Once the ego-car is outside the track or collide with other cars, 
it is counted as a collision.

\begin{figure}[!t] \centering
	\includegraphics[width=0.85\columnwidth]{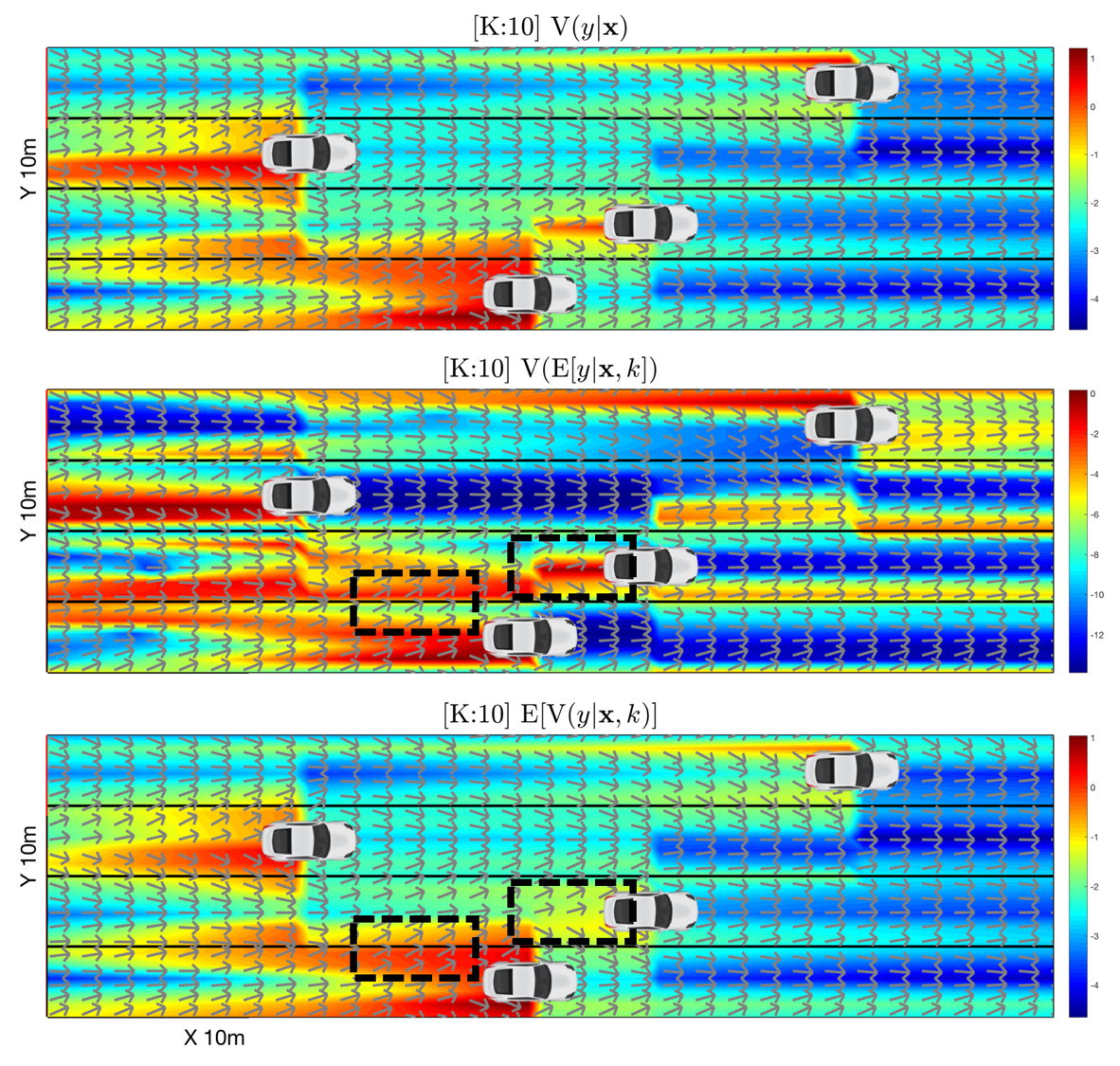}
	\label{fig:trackres_a}
	\caption{
		Different uncertainty measures on
		tracks. 
		}
	\label{fig:trackres}
\end{figure}

\begin{figure}[!t] \centering
	\includegraphics[width=0.7\columnwidth]{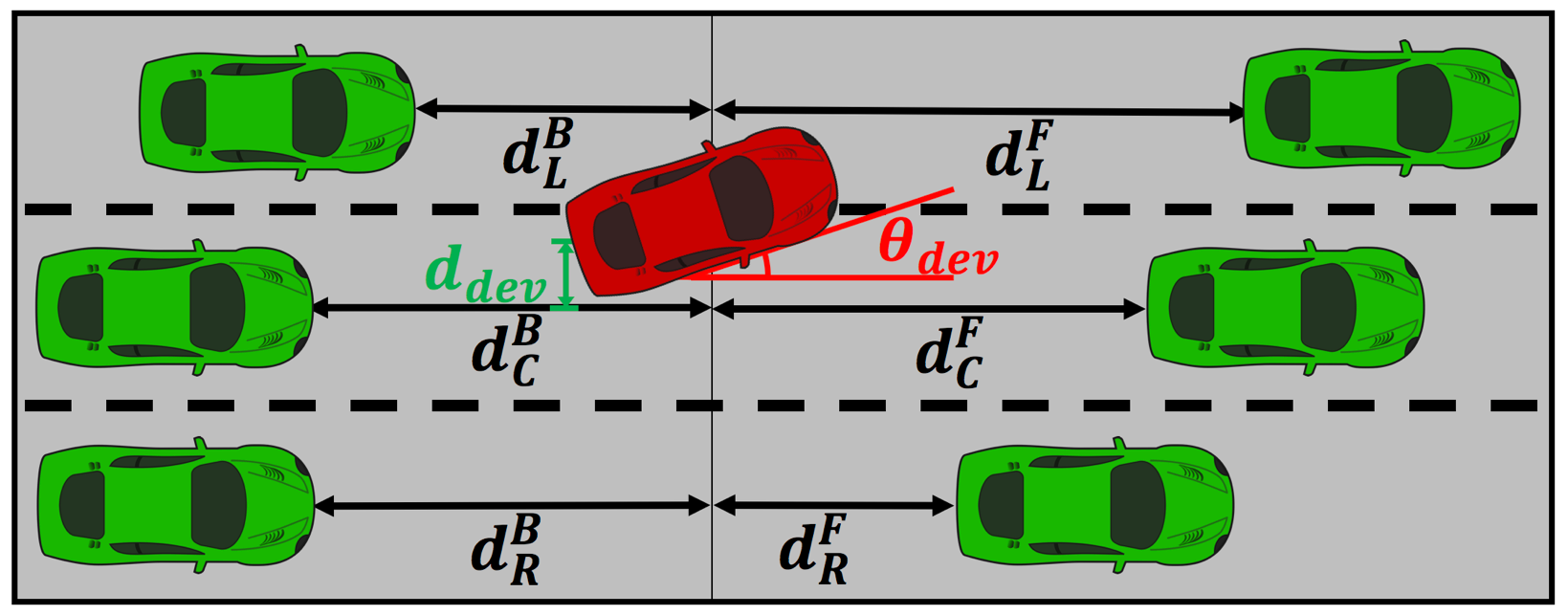}
	\caption{
		Feature descriptions for the track driving scenarios. 
		}
	\label{fig:trackfeat}
\end{figure}

To efficiently collect a sufficient number of driving demonstrations, 
we use density matching reward learning \cite{SJChoi_16_DMRL}
to automatically generate driving demonstrations 
in diverse environments by randomly changing
the initial position of an ego car 
and configurations of other cars. 
Demonstrations with collisions are excluded from the dataset
to form collision-free trajectories. 
However, it is also possible to manually collect an efficient number
of demonstrations using human-in-the loop simulations.

We define a learning-based driving policy $\pi$ 
as a mapping from input features to the trigonometric encoded
desired heading angle of a car, 
$(\cos\theta_{dev} , \, \sin \theta_{dev})$.
Figure \ref{fig:trackfeat} illustrates obtainable features of the
track driving simulator.
A seven-dimensional frontal and rearward feature representation, 
which consists of three frontal distances to the closest cars in left, center,
and right lanes in the front, 
three rearward distances to the closest cars in left, center, and right lanes in the back,
and lane deviate distance,
$(d^F_L, \, d^F_C, \, d^F_R, d^B_L, \, d^B_C, \, d^B_R, \, d_{dev})$,
are used as the input representation. 

\begin{figure*}[!t] \centering
	\includegraphics[width=1.9\columnwidth]{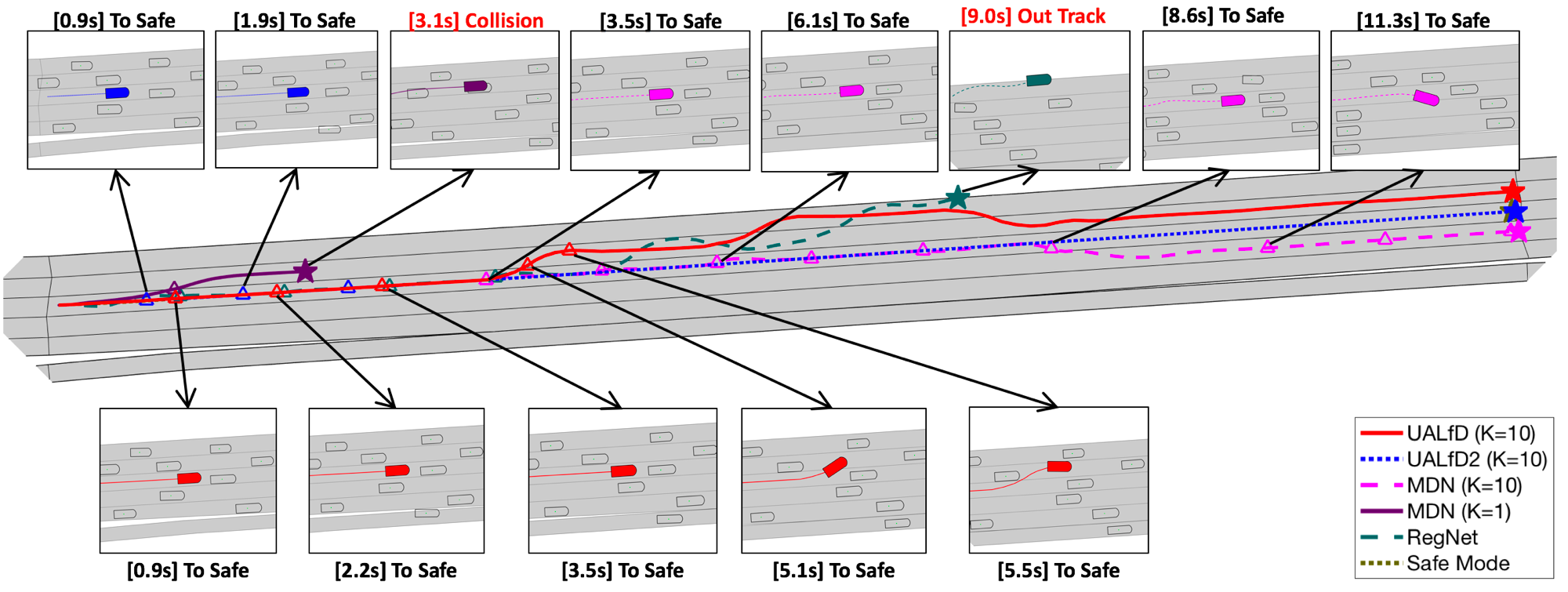}
	\caption{
		Snapshots of driving results of different LfD methods.  
		}
	\label{fig:ngsimres}
\end{figure*}

\begin{table*}[t!]  \center 
\begin{tabular}{ l | c c c c c c  } 
                    & Collision Ratio [\%] & Min. Dist. to Cars & Lane Dev. Dist. [mm] & Lane Dev. Deg. 
                    & Elapsed Time [s] & Num. Lane Change \\
	\hline \hline
	\textit{UALfD (K=10)}		& $\mathbf{0.00}$ 	& $3.67$ 			& $154.29$
						& $2.13$ 			& $15.83$ 		& $1.62$
	\\ 
	\textit{UALfD2 (K=10)}	& $\mathbf{0.00}$ 	& $6.48$ 			& $75.56$
						& $0.94$ 			& $17.47$ 		& $0.09$
	\\
	\textit{MDN (K=10)}     	& $7.55$ 			& $3.43$ 			& $289.01$
						& $3.41$ 			& $16.40$ 		& $1.67$
	\\
	\textit{MDN (K=1)}         	& $15.09$ 		& $3.34$ 			& $373.03$
						& $3.94$ 			& $17.22$ 		& $1.38$
	\\
	\textit{RegNet}        		& $28.30$ 		& $3.49$ 			& $393.48$
						& $4.86$ 			& $17.42$ 		& $2.05$
	\\	
	\textit{Safe Mode}     		& $0.00$ 			& $13.59$ 		& $5.78$
						& $0.06$ 			& $18.46$ 		& $0.00$
	\\	
\end{tabular}
\caption{Quantitative driving results in NGSIM dataset.}
\label{tbl:ngsim_full}

\begin{tabular}{ l | c c c c c c  } 
                    & Collision Ratio [\%] & Min. Dist. to Cars & Lane Dev. Dist. [mm] & Lane Dev. Deg. 
                    & Elapsed Time [s] & Num. Lane Change \\
	\hline \hline
	\textit{UALfD (K=10)}		& $\mathbf{0.00}$ 	& $4.28$ 			& $197.03$
						& $2.80$ 			& $15.54$ 		& $2.52$
	\\ 
	\textit{UALfD2 (K=10)}	& $1.16$ 			& $6.68$ 			& $113.38$
						& $1.43$ 			& $18.26$ 		& $0.04$
	\\
	\textit{MDN (K=10)}     	& $3.49$ 			& $4.15$ 			& $286.07$
						& $3.54$ 			& $15.60$ 		& $2.05$
	\\
	\textit{MDN (K=1)}         	& $25.58$ 		& $4.05$ 			& $368.23$
						& $3.91$ 			& $16.19$ 		& $2.42$
	\\
	\textit{RegNet}        		& $41.86$ 		& $4.49$ 			& $403.08$
						& $5.01$ 			& $17.79$ 		& $2.52$
	\\	
	\textit{Safe Mode}     		& $1.16$ 			& $23.57$ 		& $5.54$
						& $0.06$ 			& $20.71$ 		& $0.00$
	\\	
\end{tabular}
\caption{Quantitative driving results in NGSIM dataset with $80\%$
of cars.}
\label{tbl:ngsim_half}
\end{table*}

We trained three different network configurations:
an MDN with ten mixtures, \textit{MDN (K=10)},
MDN with one mixture, \textit{MDN (k=1)}\footnote{
This is identical to the density network used in
\cite{Kendall_17, Lakshminarayanan_16}
},
and a baseline fully connected layer, \textit{RegNet}, 
trained with a squared loss. 
All networks have two hidden layers with $256$ nodes
where a $tanh$ activation function is used. 
The proposed 
uncertainty-aware learning from demonstration method, \textit{UALfD},
switches its mode to the safe policy when the
$\log$ of \textit{explained} variance is higher that $-2$. 
As the variance is not scaled, we manually tune the threshold. 
However, we can chose the threshold based on the 
percentile of an empirical cumulative probability distribution
similar to \cite{Richter_17}. 
We also implemented uncertainty-aware learning from 
demonstration method, \textit{UALfD2}
which utilizes $\mathrm{E}[\mathrm{V}(y|\Bx, k)]$
instead of 
$\mathrm{V}(\mathrm{E}[y|\Bx, k])$
to justify the usage of using 
$\mathrm{V}(\mathrm{E}[y|\Bx, k])$
as an uncertainty measure of LfD. 
To avoid an immediate collision, \textit{UALfD} switches to 
the \textit{safe} mode when $d^F_C$ is below $1.5\,m$
and other methods switches when 
$d^F_C$ is below $2.5\,m$.\footnote{
We also tested with changing the distance threshold to $1\,m$,
but the results were worse than $2.5\,m$ in terms of collision ratio. 
}

Figure \ref{fig:trackres} illustrates 
total variance 
$\mathrm{V}(y | \Bx)$, 
\text{explainable} variance 
$\mathrm{V}(\mathrm{E}[y|\Bx, k])$, and
\textit{unexplained} varaince
$\mathrm{E}[\mathrm{V}(y|\Bx, k)]$
estimated using an MDN with ten mixtures
at each location. 
Here, we can see clear differences between 
$\mathrm{V}(\mathrm{E}[y|\Bx, k])$
and 
$\mathrm{E}[\mathrm{V}(y|\Bx, k)]$
in two squares with black dotted lines
where $\mathrm{V}(\mathrm{E}[y|\Bx, k])$
can better capture model uncertainty 
possibly due to the lack of training data. 
Furthermore, we can see that the regions 
where the desirable heading depicted with gray arrows
are not accurate, e.g., leading to a collision, has
higher variance and it does not necessarily depend on the 
distance between the frontal car. 
This supports that our claim in that
$\mathrm{V}(\mathrm{E}[y|\Bx, k])$
is more suitable for estimating modeling error.

Once a desired heading is achieved, 
the ego-car is controlled at $10$Hz using a simple feedback controller
for an angular velocity $w$ with 
$w = \min(2 \cdot \text{sign}(\theta_{diff}) \cdot \theta_{diff}^2, w_{max}) $
where $\theta_{diff}$ is the difference between current heading
and desired heading from the learned controller
normalized between $-180$ to $+180 degree$.
A directional velocity is fixed at $90\,$km/h and the control
frequency is set to $10\,$Hz. 
While we use a simple unicycle dynamic model, more complex dynamic 
models, e.g., vehicle and bicycle dynamic models, can be also used. 

We also carefully designed a rule-based safe controller 
that safely keeps its lane without a collision
where the directional and angular velocities are computed 
by following rules:
\begin{align*}
	v &= 
	\begin{cases}
    		v^F_C-5\,\text{m/s},		& \text{if  $d^F_C < 3$} 
		\\
    		v^R_C+3\,\text{m/s},           & \text{if  $d^R_C < 3$} 
		\\
    		\frac{v^F_C+v^R_C}{2},              & \text{otherwise}
	\end{cases}
	\\
	w &= -5*\theta_{dev} + 50*d_{dev}
\end{align*}
where $v^F_C$ and $v^R_C$ 
are the directional velocities of the frontal and rearward cars, 
respectively.

We conducted two sets of experiments, 
one with using the entire cars
and the other with using $70\%$ of the cars.
The quantitative results are shown in Table \ref{tbl:ngsim_full}
and \ref{tbl:ngsim_half}. 
The results show that the driving policy that incorporates
the proposed uncertainty measure clearly improves
both safety and stability of driving.
We would like to emphasize that the average elapsed time of 
the propose \textit{UALfD} is the shortest among 
the compared method.
Figure \ref{fig:ngsimres} shows trajectories and some snapshots of
driving results of different methods. 
Among compared methods, the proposed \textit{UALfD}, 
\textit{UALfD2}, 
\textit{MDN (K=10)},
and \textit{Safe Mode} safely navigates without colliding
with other moving cars.
On the other hand, the cars controlled with 
\textit{MDN (K=1)} and \textit{RegNet} 
collide with another car and 
move outside the track
which clearly shows the advantage of using a mixture density network 
for modeling human demonstrations.  
Furthermore, while both $\textit{UALfD}$ and $\textit{UALfD2}$
navigate without a collision, the average elapsed time 
and the average number lane changes varies greatly.
This is mainly due to the fact that 
$\mathrm{E}[\mathrm{V}(y|\Bx, k)]$
captures the measure noise rather than 
the model uncertainty which makes the control 
conservative similar to that of the \textit{Safe Mode}.

\section{Conclusion}

In this paper, we proposed a novel uncertainty estimation method
using a mixture density network. 
Unlike existing approaches that rely on ensemble of multiple models
or Monte Carlo sampling with stochastic forward paths, 
the proposed uncertainty acquisition method can run with 
a single feedforward model without computationally-heavy sampling. 
We show that the proposed uncertainty measure can be decomposed
into \textit{explained} and \textit{unexplained} variances 
and analyze the properties with three different cases: 
\textit{absence of data}, \textit{heavy measurement noise}, 
and \textit{composition of functions} scenarios and
show that it can effectively distinguish the three cases using 
the two types of variances.
Furthermore, we propose an uncertainty-aware 
learning from demonstration method
using the proposed uncertainty estimation and 
successfully applied to real-world driving dataset. 

\bibliographystyle{IEEEtran}
\bibliography{references}

\begin{thebibliography}{10}
\providecommand{\url}[1]{#1}
\csname url@samestyle\endcsname
\providecommand{\newblock}{\relax}
\providecommand{\bibinfo}[2]{#2}
\providecommand{\BIBentrySTDinterwordspacing}{\spaceskip=0pt\relax}
\providecommand{\BIBentryALTinterwordstretchfactor}{4}
\providecommand{\BIBentryALTinterwordspacing}{\spaceskip=\fontdimen2\font plus
\BIBentryALTinterwordstretchfactor\fontdimen3\font minus
  \fontdimen4\font\relax}
\providecommand{\BIBforeignlanguage}[2]{{%
\expandafter\ifx\csname l@#1\endcsname\relax
\typeout{** WARNING: IEEEtran.bst: No hyphenation pattern has been}%
\typeout{** loaded for the language `#1'. Using the pattern for}%
\typeout{** the default language instead.}%
\else
\language=\csname l@#1\endcsname
\fi
#2}}
\providecommand{\BIBdecl}{\relax}
\BIBdecl

\bibitem{He_16}
K.~He, X.~Zhang, S.~Ren, and J.~Sun, ``Deep residual learning for image
  recognition,'' in \emph{Proc. of the IEEE conference on Computer Vision and
  Pattern Recognition}, 2016, pp. 770--778.

\bibitem{Collobert_08}
R.~Collobert and J.~Weston, ``A unified architecture for natural language
  processing: Deep neural networks with multitask learning,'' in \emph{Proc. of
  the International Conference on Machine Learning}, 2008, pp. 160--167.

\bibitem{Schulman_15}
J.~Schulman, S.~Levine, P.~Abbeel, M.~I. Jordan, and P.~Moritz, ``Trust region
  policy optimization.'' in \emph{Proc. of the International Conference on
  Machine Learing}, 2015, pp. 1889--1897.

\bibitem{Amodei_16}
D.~Amodei, C.~Olah, J.~Steinhardt, P.~Christiano, J.~Schulman, and D.~Man{\'e},
  ``Concrete problems in ai safety,'' \emph{arXiv preprint arXiv:1606.06565},
  2016.

\bibitem{Tesla_16}
\BIBentryALTinterwordspacing
AP and REUTERS, ``Tesla working on 'improvements' to its autopilot radar
  changes after model s owner became the first self-driving fatality.'' June
  2016. [Online]. Available: \url{https://goo.gl/XkzzQd}
\BIBentrySTDinterwordspacing

\bibitem{Kendall_17}
A.~Kendall and Y.~Gal, ``What uncertainties do we need in {B}ayesian deep
  learning for computer vision?'' \emph{arXiv preprint arXiv:1703.04977}, 2017.

\bibitem{Bishop_94}
C.~M. Bishop, ``Mixture density networks,'' 1994.

\bibitem{Brando_17}
A.~Brando~Guillaumes, ``Mixture density networks for distribution and
  uncertainty estimation,'' Master's thesis, Universitat Polit{\`e}cnica de
  Catalunya, 2017.

\bibitem{Lakshminarayanan_16}
B.~Lakshminarayanan, A.~Pritzel, and C.~Blundell, ``Simple and scalable
  predictive uncertainty estimation using deep ensembles,'' \emph{arXiv
  preprint arXiv:1612.01474}, 2016.

\bibitem{Gal_16}
Y.~Gal and Z.~Ghahramani, ``Dropout as a {B}ayesian approximation: Representing
  model uncertainty in deep learning,'' in \emph{Proc. of the International
  Conference on Machine Learing}, 2016, pp. 1050--1059.

\bibitem{Gal_16_thesis}
Y.~Gal, ``Uncertainty in deep learning,'' Ph.D. dissertation, PhD thesis,
  University of Cambridge, 2016.

\bibitem{Mclachlan_88}
G.~J. McLachlan and K.~E. Basford, \emph{Mixture models: Inference and
  applications to clustering}.\hskip 1em plus 0.5em minus 0.4em\relax Marcel
  Dekker, 1988, vol.~84.

\bibitem{Colyar_07}
J.~Colyar and J.~Halkias, ``Us highway 101 dataset,'' Federal Highway
  Administration (FHWA), Tech. Rep., 2007.

\bibitem{Ross_13}
S.~Ross, ``Interactive learning for sequential decisions and predictions,''
  Ph.D. dissertation, Carnegie Mellon University, 2013.

\bibitem{Srivastava_14}
N.~Srivastava, G.~E. Hinton, A.~Krizhevsky, I.~Sutskever, and R.~Salakhutdinov,
  ``Dropout: a simple way to prevent neural networks from overfitting.''
  \emph{Journal of machine learning research}, vol.~15, no.~1, pp. 1929--1958,
  2014.

\bibitem{Kahn_17}
G.~Kahn, A.~Villaflor, V.~Pong, P.~Abbeel, and S.~Levine, ``Uncertainty-aware
  reinforcement learning for collision avoidance,'' \emph{arXiv preprint
  arXiv:1702.01182}, 2017.

\bibitem{Richter_17}
C.~Richter and N.~Roy, ``Safe visual navigation via deep learning and novelty
  detection,'' in \emph{Proc. of the Robotics: Science and Systems Conference},
  2017.

\bibitem{Shazeer_17}
N.~Shazeer, A.~Mirhoseini, K.~Maziarz, A.~Davis, Q.~Le, G.~Hinton, and J.~Dean,
  ``Outrageously large neural networks: The sparsely-gated mixture-of-experts
  layer,'' \emph{arXiv preprint arXiv:1701.06538}, 2017.

\bibitem{Duda_73}
R.~O. Duda, P.~E. Hart, and D.~G. Stork, \emph{Pattern classification}.\hskip
  1em plus 0.5em minus 0.4em\relax Wiley, New York, 1973.

\bibitem{Abadi_16}
M.~Abadi, A.~Agarwal, P.~Barham, E.~Brevdo, Z.~Chen, C.~Citro, G.~S. Corrado,
  A.~Davis, J.~Dean, M.~Devin \emph{et~al.}, ``Tensorflow: Large-scale machine
  learning on heterogeneous distributed systems,'' \emph{arXiv preprint
  arXiv:1603.04467}, 2016.

\bibitem{SJChoi_16_DMRL}
S.~Choi, K.~Lee, A.~Park, and S.~Oh, ``Density matching reward learning,''
  \emph{arXiv preprint arXiv:1608.03694}, 2016.

\end{thebibliography}

\end{document}